\documentclass[journal]{IEEEtran}

\usepackage{verbatim}
\usepackage{amsfonts}
\usepackage{amssymb}
\usepackage{stfloats}
\usepackage{cite}
\usepackage{graphicx}
\usepackage{psfrag}
\usepackage{amsmath}
\usepackage{array}
\usepackage{epstopdf}
\usepackage{authblk}
\usepackage{graphicx} 
\usepackage{amsthm} 
\usepackage{lipsum}
\usepackage{verbatim} 
\usepackage{authblk}
\usepackage{mathtools}
\usepackage{cuted}
\usepackage{booktabs}

\usepackage{amsmath}
\usepackage{mathrsfs}

\usepackage{algorithmic}

\usepackage{array}

\ifCLASSOPTIONcompsoc
\usepackage[caption=false,font=normalsize,labelfont=sf,textfont=sf]{subfig}
\else
\usepackage[caption=false,font=footnotesize]{subfig}
\fi

\usepackage{url}
\usepackage{graphicx,amsmath,amssymb,amsfonts}
\usepackage{algorithmic,algorithm}

\usepackage{hyperref}
\hypersetup{colorlinks=true, citecolor=blue}

\usepackage{setspace}

\makeatletter

\newcommand{\Rmnum}[1]{\expandafter\@slowromancap\romannumeral #1@}
\makeatother

\usepackage{framed} 

\usepackage{cancel}

\usepackage{booktabs}
\usepackage{tabularx,makecell,multirow}

\usepackage[table]{xcolor}
\usepackage{pifont} 

\usepackage{diagbox}

\usepackage{multirow}

\begin{document}
	\bstctlcite{ref:BSTcontrol}

	\title{\fontsize{22.5 pt}{\baselineskip}\selectfont Multi-Task Semantic Communications via Large Models}
	
	\author{Wanli Ni, Zhijin Qin, Haofeng Sun, Xiaoming Tao, and Zhu Han
	
	\vspace{-6 mm}
		\thanks{Wanli Ni, Zhijin Qin, and Xiaoming Tao are with the Department of Electronic Engineering, Tsinghua University, Beijing 100084, China, with the State Key Laboratory of Space Network and Communications, Beijing 100084, China, and also with Beijing National Research Center for Information Science and Technology, Beijing 100084, China (email: niwanli@tsinghua.edu.cn, qinzhijin@tsinghua.edu.cn, taoxm@tsinghua.edu.cn).}
		\thanks{Haofeng Sun is with the State Key Laboratory of Networking and Switching Technology, Beijing University of Posts and Telecommunications, Beijing 100876, China (email: sunhaofeng@bupt.edu.cn).}
		\thanks{Zhu Han is with the Electrical and Computer Engineering, University of Houston, Houston TX 77004, USA (e-mail: hanzhu22@gmail.com).}
	}

	\maketitle
	
	\begin{abstract}
		Artificial intelligence (AI) promises to revolutionize the design, optimization and management of next-generation communication {systems}.
		In this article, we explore the integration of large AI models (LAMs) into semantic communications (SemCom) by leveraging their multi-modal data processing and generation capabilities.
		Although LAMs bring unprecedented abilities to extract semantics from raw data, this integration entails multifaceted challenges including {high resource demands}, model complexity, and {the need for} adaptability across {diverse} modalities and tasks.
		To overcome these challenges, we propose a LAM-based multi-task SemCom (MTSC) architecture, which includes an adaptive model compression strategy and a federated split fine-tuning {approach to facilitate} the efficient deployment of LAM-based semantic models in {resource-limited} networks.
		Furthermore, a retrieval-augmented generation scheme is {implemented to synthesize the most recent} local and global knowledge bases to {enhance} the accuracy of semantic extraction and content generation, {thereby} improving the inference performance.
		Finally, simulation results {demonstrate the efficacy} of the proposed LAM-based MTSC architecture, highlighting the performance {enhancements across various} downstream tasks under varying channel conditions.
	\end{abstract}
	
	
	\vspace{-2 mm}
	\section{Introduction}
	The advent of 6G networks promises a new era in wireless communication, characterized by extreme connectivity, ubiquitous intelligence, and service capabilities.
	In future 6G era, the explosion of multi-modal data and the increasing demand for intelligent services propel the quest for groundbreaking innovations, which inspire us to rethink communication systems \cite{Qin2024AI}.
	Among all 6G candidate technologies, semantic communication (SemCom) heralds a new frontier that promises to transcend the limitations of conventional bit transmission, focusing on the conveyance of meaning and feature of data rather than mere sequences of symbols \cite{Yang2023Semantic}.
	This paradigm shift is not only a simple step towards integrating artificial intelligence (AI) into the fabric of communication, but also reshape how we understand and utilize communication systems \cite{Qin2024AI}.
	Overall, in addition to being solely a passage for data flow, SemCom also acts as a key enabling technology for future smart scenarios having a number of intelligent applications \cite{Yang2023Semantic}.
	
	{
	In prior works, uni-modal SemCom focuses on optimizing the transmission and reception of data within a single modality.
	For text tasks, uni-modal SemCom is used to compress messages based on their semantic meaning rather than their literal representation \cite{Qin2024AI}.
	It reduces bandwidth requirements by disregarding syntactical details or redundant information.
	Similarly, for audio tasks, uni-modal SemCom aims to transmit the semantic content of speech signals accurately \cite{Weng2023Deep}.
	By encoding spoken language semantically, uni-modal SemCom improves noise resilience by reducing the impact of background noise or distortions on the transmitted audio.
	For visual tasks, uni-modal SemCom mainly identifies and encodes key objects, scenes, or features within the visual data, discarding less relevant information \cite{Bourtsoulatze2019Deep}.
	Recently, by extracting semantics from different data modalities, multi-modal SemCom is able to achieve a more comprehensive understanding of the multi-modal input \cite{Xie2022Task}.
	For instance, in a video conferencing application, multi-modal SemCom systems first predict the importance of video segments based on audio semantics, and then allocate more resources to transmit crucial audio and video semantics \cite{Jiang2023Video}.
	Therefore, in multi-modal SemCom, by cooperatively encoding the semantic information of audio and video streams, the data load is reduced and the overall transmission quality is improved \cite{Barbarossa2023Semantic}.
	In the following, three major challenges faced by existing SemCom systems are summarized, which serve as the motivations for this work.}
		
	\textbf{Challenge 1: Inadequate adaptability to multi-modality and multi-tasking:}
	Many current SemCom systems are limited by their inability to handle both multi-modality and multi-tasking, concentrating on single-source transmissions (e.g., image or text).
	However, real-world applications frequently involve handling information that merges multiple modalities for diverse concurrent tasks, including but not limited to speech recognition, image classification, and linguistic analysis.
	To cope with these challenges, one opportunity is to design a generalized model architecture to enable effective multi-modal fusion, thereby boosting the efficiency and accuracy of multi-modal data processing.
	Equally important is the development of flexible training strategies that can manage multiple tasks at once, aligning with the rising tide of varied user needs.
	
	\textbf{Challenge 2: Model deployment and fine-tuning in resource-limited networks:}
	At the network edge, hardware provisions of clients are starkly restricted.
	This gives rise to an immediate challenge: how to ensure the efficient operation of sizable semantic models amidst the stringent resource limitations imposed by edge devices.
	To mitigate this issue, model compression becomes imperative to diminish computational intensity and curtail memory occupation.
	Moreover, some issues such as limited bandwidth resources, private risk and high latency when transmitting local datasets to the server make the traditional centralized training impractical.
	Therefore, it is particularly important to study lightweight deployment and distributed fine-tuning techniques.
	This can effectively reduce the delay and bandwidth pressure of data transmission, and also ensure that user privacy is not disclosed during the collaborative fine-tuning.

	\textbf{Challenge 3: Inaccurate content generation using outdated knowledge bases:}	
	SemCom systems rely on rich knowledge bases to accurately capture the {semantics of user inputs}.
	However, {existing} SemCom {systems face challenges in swiftly integrating} the {most current} semantics due to {model updating delay}.  
	This {limitation adversely affects} system performance, {particularly} when addressing tasks {related to recent} events or hot topics.
	{Additionally}, the {lack} of domain-specific knowledge {restricts the ability} of SemCom models to {seamlessly incorporate} external information into their outputs, {which is essential for generating personalized content that meets} individual user preferences.
	To address these {challenges, it is crucial to develop a plug-and-play mechanism for updating knowledge bases}, {thereby} ensuring {that} the system remains {aligned} with the most up-to-date semantic landscape.

	To tackle the aforementioned challenges, in this article, we propose a large AI model (LAM)-based multi-task SemCom (MTSC) architecture, which is capable of handling various data modalities and {performing} multiple tasks {concurrently}.
	To facilitate the implementation of this architecture, several key techniques are proposed. {Firstly}, an adaptive model compression {strategy is developed to achieve} lightweight deployment.
	{Secondly}, a federated split fine-tuning method is {introduced to allow for} efficient parameter updates, {thereby enabling} rapid task adaptation while {preserving} user privacy. 
	{Furthermore}, a retrieval-augmented generation (RAG) {approach is employed to improve} the accuracy of semantic extraction and content generation by retrieving relevant information from plug-in knowledge bases. 
	{Finally}, an importance-aware semantic transmission technique is proposed to dynamically adjust the encoding rate {based on semantic importance} and real-time channel conditions, thus enhancing the efficiency and reliability of data transmission.
		
	The rest of this article is organized as follows. 
	Section II first provides a brief overview of large models for language tasks and beyond.
	In Section III, the LAM-based MTSC architecture and key techniques of model deployment and fine-tuning are detailed.
	Potential application scenarios of the proposed MTSC are introduced in Section IV.
	Section V concludes this article with several future directions.

	\section{An Overview of Large Models for Language Tasks and Beyond}

	\subsection{LAM for Language Tasks}
	The {emergence} of LAMs has {significantly transformed the domain} of language processing tasks \cite{Humza2023Overview}.
	As the {scale} and complexity of AI models continue to {expand}, we are witnessing the {progression} of new frontiers ranging from language {comprehension} to content generation \cite{Liang2024Generative}.
	{In particular}, LAMs, {exemplified} by architectures such as Transformers and their derivatives, have {exhibited} an {unparalleled capability} in understanding the intricate {subtleties} of human language \cite{Humza2023Overview}. 
	{These models} excel in semantic parsing, {allowing} them to {grasp} not only the explicit meanings of words and phrases but also the implicit relationships, idiomatic expressions, and emotional {undertones present} in the text.
	{Consequently, applications powered by these LAMs, such as automated customer service chatbots, are now engage in more natural and empathetic dialogues, effectively responding to customer inquiries and emotional cues.}

	\subsection{LAM for Tasks Beyond Language}
	Going beyond language tasks, LAMs are increasingly {utilized in the domains of} computer vision, audio processing, and {task-oriented SemCom systems} \cite{Quan2025Large, shukor2023unival, Zhao2024LaMoSC}.
	{
	Their advanced representation learning capabilities enable LAMs to effectively extract significant features from various data modalities, leading to state-of-the-art performance on a wide range of tasks.
	For example, the authors of \cite{shukor2023unival} proposed UnIVAL, a unified model designed to integrate various modalities, representing a significant advancement towards general AI. 
	Furthermore, to improve the visual reconstruction quality in conventional SemCom systems, the authors of \cite{Zhao2024LaMoSC} developed a SemCom system driven by LAM that utilizes multi-modal features for the recovery of visual information.
	Additionally, a LAM-based multi-modal SemCom system was proposed in \cite{Jiang2024Multimodal}, where textual data was used to achieve multi-modal alignment and maintain semantic consistency.}
	
	\vspace{-2 mm}
	\section{Proposed LAM-Based MTSC Architecture}
	 Existing SemCom systems face challenges, such as weak multi-tasking adaptability, limited client resources, and outdated knowledge bases.
	 {To address these issues, we propose a LAM-based MTSC architecture, along with a lightweight strategy, a fine-tuning method, and a RAG scheme to develop intelligent communication systems.}
	
	\vspace{-2 mm}
	\subsection{Architecture Design of LAM-Based MTSC}
\begin{figure*}[t]
		\centering
		\includegraphics[width=7 in]{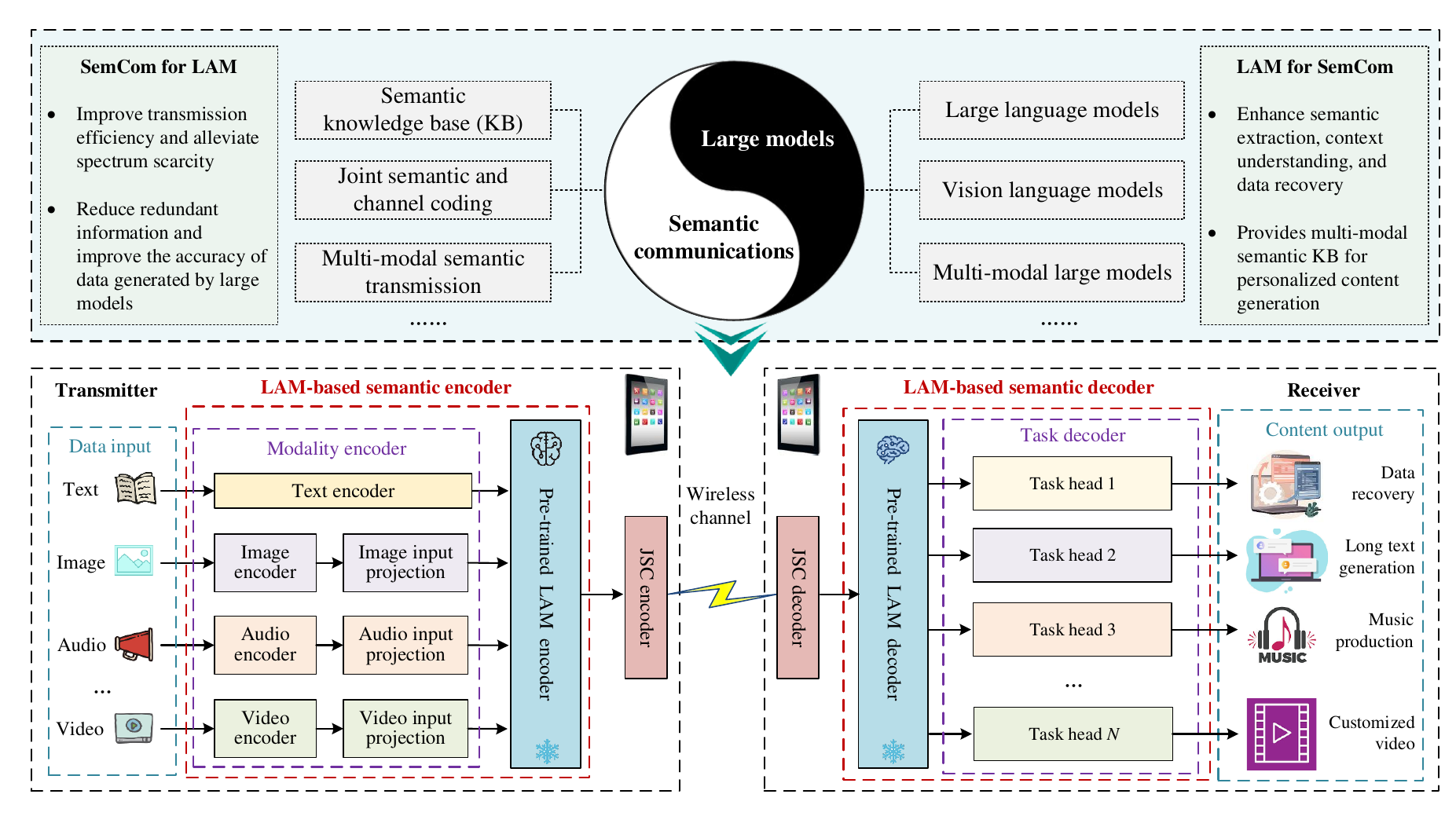}
		\caption{An illustration of the proposed LAM-based MTSC architecture. The transmitter consists of modality encoders, a pre-trained LAM encoder, and a JSC encoder. The receiver comprises a JSC decoder, a pre-trained LAM decoder, and task decoders.}
		\label{Fig1}
	\end{figure*}

	The complexity and heterogeneity of multi-modal data pose significant challenges for unified semantic extraction and fusion. 
	Traditional SemCom systems often struggle to effectively capture and integrate the semantic information embedded within diverse data modalities \cite{Jiang2024Multimodal}.
	To address these limitations, a LAM-based MTSC architecture is proposed for achieving efficient semantic alignment and parsing across multi-modal data, thereby enhancing the accuracy and efficiency of multi-task SemCom.
	The proposed architecture is shown in Fig. \ref{Fig1}, and its main components are given as follows.
	
	\subsubsection{LAM-Based Semantic Encoder}
	{
	At the transmitter, modality-specific encoders and input projections are combined with the pre-trained LAM encoder to function as a comprehensive multi-modal semantic encoder.
	The key components of the LAM-based semantic encoder are detailed below.}
		
	\begin{itemize}
		\item 
		\textbf{Modality encoder:}
		{
		These encoders are {designed to handle specific} data modalities.
		For {example}, convolutional neural networks (CNNs) {are utilized} for {the} encoding {of images}, {whereas} recurrent neural {networks (RNNs)} or Transformers are applied for {the encoding of} text and audio.
		Each encoder captures the unique features of its respective modality, {thereby converting} raw data into high-level semantic representations.}

		\item
		\textbf{Pre-trained LAM encoder:}
		{
		The pre-trained LAM encoder functions as the fusion center, utilizing large models to understand and generate contextually relevant content. This encoder receives the semantic representations from the modality encoder as input and maps them into a unified semantic space. This fusion process not only captures the intrinsic semantic relationships among different modalities, but also enhances the overall semantic richness of the encoded data.}
	\end{itemize}
		
	\subsubsection{JSC Encoder and Decoder}
	The joint source-channel (JSC) encoder and decoder play a role in optimizing the transmission of semantic information over wireless channels~\cite{Bourtsoulatze2019Deep}.
	
	\begin{itemize}
		\item 
		\textbf{JSC encoder:} 
		At the transmitter, the JSC encoder dynamically adapts its encoding strategy based on real-time channel conditions and the characteristics of the semantic information. It employs variable-rate coding techniques to balance the trade-off between data rate and transmission reliability, ensuring that the most relevant and informative semantic content is transmitted efficiently. This adaptive approach significantly improves the robustness of the system against channel impairments and variability.
	
		\item 
		\textbf{JSC decoder:} 
		At the receiver, the JSC decoder reconstructs the semantic information from the received signals. It leverages advanced decoding algorithms that are capable of recovering the original semantic content with high fidelity, even in the presence of noise and distortions. By closely coordinating with the encoder, the JSC decoder ensures that the reconstructed semantic information accurately reflects the original intent and context.
	\end{itemize}

	\subsubsection{LAM-Based Semantic Decoder}
	{
	At the receiver, the LAM-based semantic decoder incorporates various task-specific decoders, thereby facilitating the concurrent execution of multiple downstream tasks.
	
	\begin{itemize}
		\item 
		\textbf{Pre-trained LAM decoder:} 
		The LAM decoder functions as an interface for the alignment and interpretation of the semantic information obtained by the JSC decoder.
		Specifically, by leveraging its embedded knowledge to evaluate both task-specific and task-general features, the LAM decoder produces task-oriented semantics, which are then fed into task decoders for either data restoration or content generation.
				
		\item
		\textbf{Task decoder:} 
		The task decoder is comprised of output projections and modality-specific decoders.
		Based on the semantics received, the task decoder orchestrates the execution of multiple different downstream tasks, such as data recovery, visual question answering (VQA), and multi-modal content generation.
		Through the collaborative reasoning of various task heads, the task decoder is able to complete multiple tasks in parallel, thereby delivering customized services to users.
	\end{itemize}
	}

	\begin{figure*}[t]
		\centering
		\includegraphics[width=6.8 in]{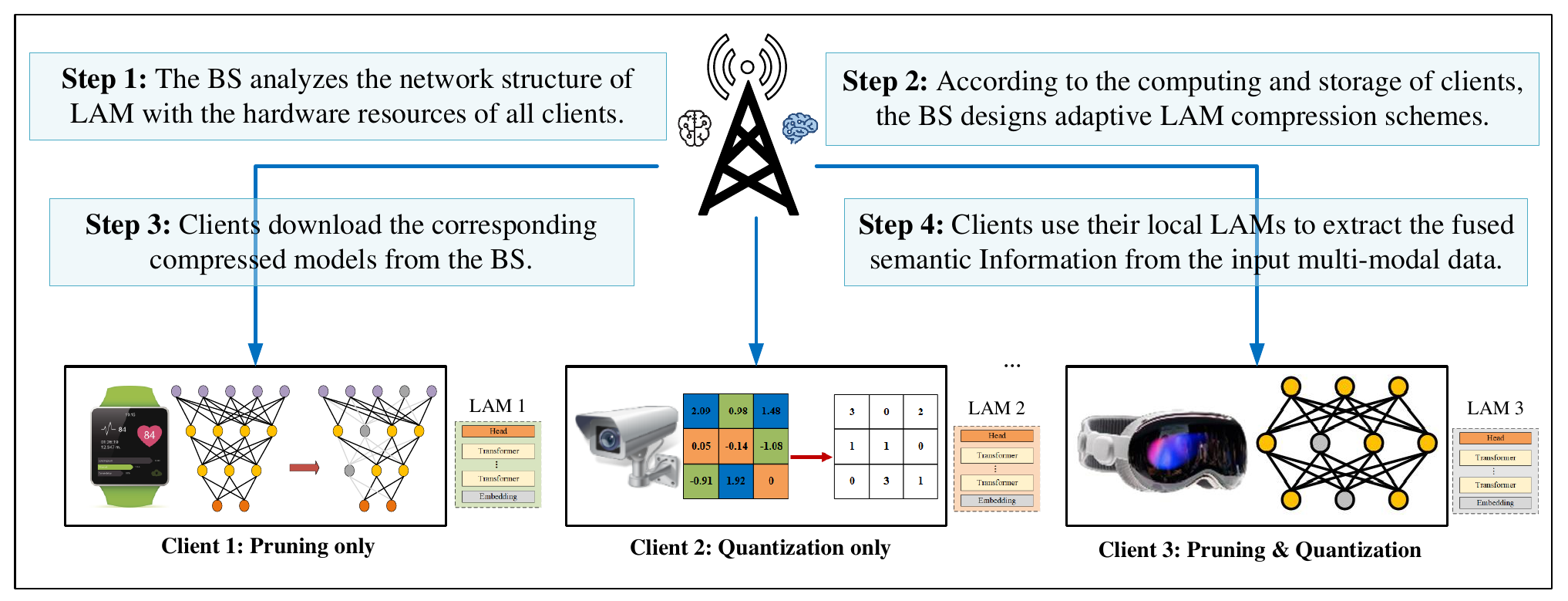}
		\caption{Adaptive model compression for lightweight deployment of LAM-based semantic models at the resource-constrained network edge.}
		\label{Fig2}
	\end{figure*}

	\subsection{Model Compression and Federated Fine-Tuning}
	
	\subsubsection{Lightweight Model Deployment at the Network Edge}
	The deployment of LAMs at the resource-constrained network edge presents unique challenges.
	To {mitigate these issues}, a {strategy for} lightweight model deployment is {developed}, {aimed at minimizing} resource {requirements} and {rendering LAMs feasible} for devices with {limited capabilities}, such as wearables {and} smartwatches.
	As shown in Fig.~\ref{Fig2}, in the first step, the base station (BS) {evaluates} the computing and storage capabilities of clients.
	{Subsequently}, in the second step, the BS designs {tailored} compression schemes for {each client} by {employing various} compression techniques \cite{Zhu2023Survey}, including model quantization, parameter pruning, and knowledge distillation, etc.
	{In general, customized compression strategies (e.g., pruning rate and quantization level) are determined by solving an optimization problem with some objective (e.g., accuracy and delay or both).}
	Next, in the third and fourth steps, clients download the respective compressed models from the BS and use them to extract multi-modal semantic Information.
	For {example}, as {depicted} in Fig.~\ref{Fig2}, Client 1 {solely applies} model pruning, Client 2 {only utilizes} post-training quantization, while Client 3 {integrates} both model pruning and quantization, {thereby} achieving an optimal {equilibrium} between performance and resource efficiency.
	{
		Furthermore, the adoption of lightweight variants or the compression of LAMs can substantially reduce inference time, thereby satisfying the demands for real-time performance.
	}

	\begin{figure}[t]
		\centering
		\includegraphics[width=3.5 in]{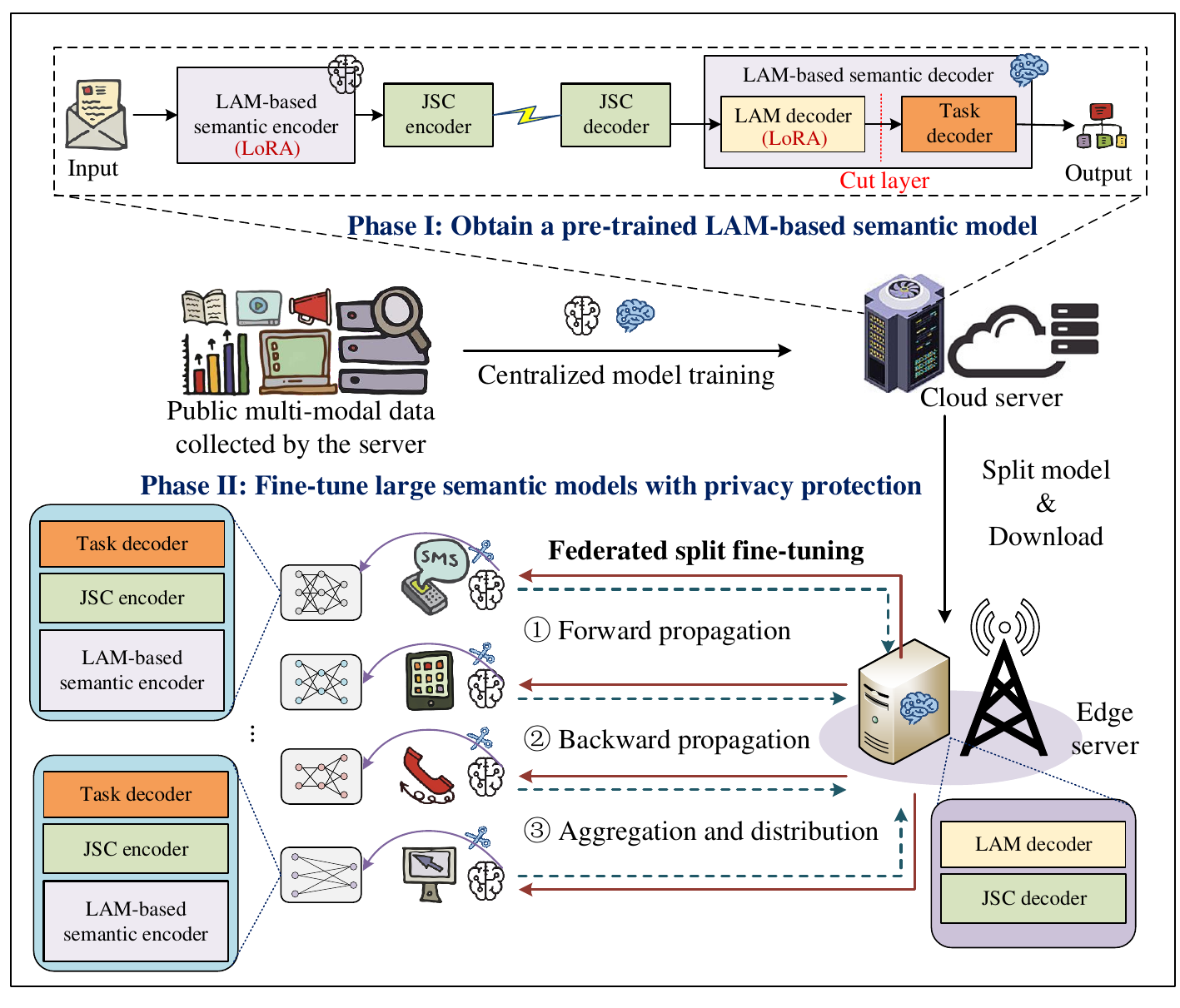}
		\caption{An illustration of the proposed federated split fine-tuning method for training large semantic models in wireless networks. During Phase I, a multi-modal semantic model is obtained by training the LAM using public multi-modal data collected by the server. In Phase II, semantic models are further fine-tuned through federated split fine-tuning.
		}
		\label{Fig3}
		\vspace{-3 mm}
	\end{figure}

	\subsubsection{Federated Split Fine-Tuning with Privacy Protection}
	
	{During} the parameter initialization phase, a LAM {utilizing} the encoder-decoder Transformer architecture is {employed}, {having been pre-trained on publicly available datasets hosted} on cloud servers.
	{In the subsequent} parameter fine-tuning phase, a federated split fine-tuning method integrated with low-rank adaptation (LoRA) and model splitting techniques is proposed.
	The pre-trained layers of the LAM remain unchanged, while trainable low-rank decomposition matrices are embedded into the existing layers of the pre-trained model.
	{
	As exemplified in Fig.~\ref{Fig3}, clients maintain a LAM-based semantic encoder, a JSC encoder, and a task decoder; meanwhile, the edge server holds a JSC decoder and a LAM decoder.
	Note that, the task decoder is split and deployed on the client, which eliminates the necessity of transmitting ground-truth data labels or task outcomes, thereby preserving user privacy.}
	The workflow of the proposed federated split fine-tuning method is detailed as follows.
	\begin{itemize}
		\item 
		\textbf{Forward propagation:}
		Clients feed local data into the LAM-based semantic encoder and the JSC encoder.
		The output from the JSC encoder is then transmitted to the edge server via a wireless channel for further processing by the JSC decoder and LAM decoder.
		Finally, the LAM decoder's output is returned to the client, where the task decoder generates the final output.
		
		\item
		\textbf{Backward propagation:}
		Clients compute the loss function to obtain gradients for the task decoder, which are then transmitted to the edge server to guide the server-side model parameter updates.
		The server, in turn, propagates its computed gradient information back to the client.
		Then, clients update their device-side models, including the modality encoder, low-rank decomposed matrices, and the JSC encoder.
		
		\item
		\textbf{Aggregation and distribution:}
		After multiple iterations of above steps, all clients send their local model updates to the edge server for aggregation.
		The aggregated model parameters are then distributed back to all clients for the subsequent training round.
	\end{itemize}

\begin{figure*}[t]
	\centering
	\includegraphics[width=7 in]{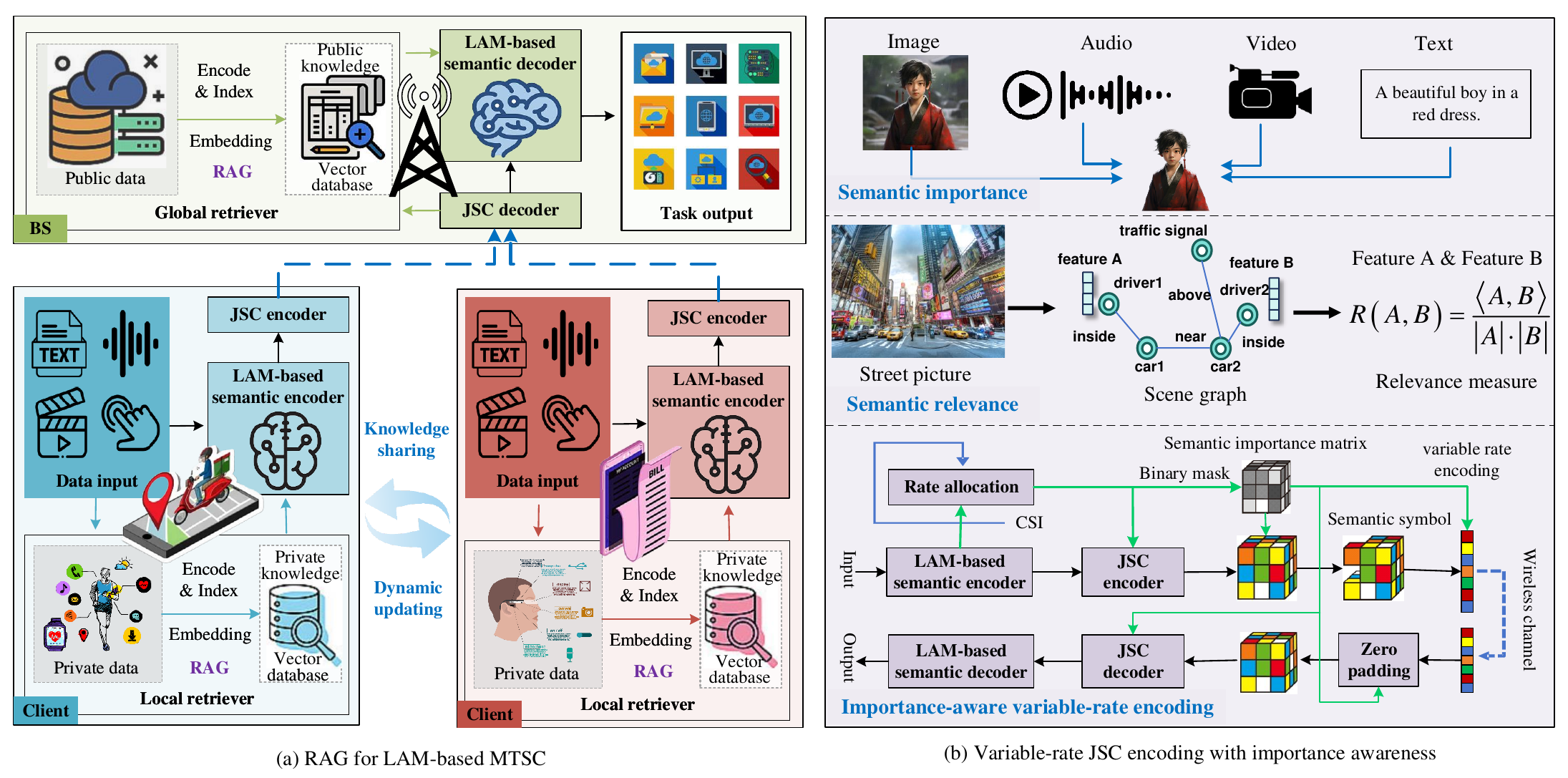}
	\caption{An illustration of RAG-enhanced LAM-based MTSC and importance-aware semantic transmission in wireless networks.}
	\label{Fig4}
\end{figure*}

	\subsection{Knowledge Base Update and Semantic Transmission}
	
	{
	\subsubsection{RAG-Based Knowledge Base Update}	
	Over time, the knowledge base of large models can become outdated. At this point, it becomes increasingly difficult for large models to accurately extract semantic information and generate reliable content.
	Although federated learning can be used to fine-tune parameters, frequently updating large models is time-consuming and energy intensive, which poses a significant challenge for practical implementation.
	To address this limitation, RAG technology can be exploited, which facilitates the rapid update of local and global knowledge bases by seamlessly integrating the latest external data sources.
	As illustrated in Fig. \ref{Fig4}(a), by transferring the private and public data to the corresponding vector database, the semantic extraction and parsing capabilities of the client and the edge server can be enhanced rapidly.
	This approach eliminates the need for model fine-tuning and effectively improves system performance in a privacy-preserving manner.
	}

	\subsubsection{Importance-Aware Semantic Transmission}
	
	The transmission of redundant data significantly {impairs} communication efficiency {and may compromise} the accuracy of SemCom. 
	{Therefore, it is imperative to assess} the importance of specific {semantic elements} and to establish a {model for evaluating} semantic relevance in multi-modal data. {In this context}, a variable-rate JSC encoding method {that integrates} semantic importance and channel conditions is proposed, as illustrated in Fig. \ref{Fig4}(b).
	This {methodology seeks} to {improve} the efficiency and reliability of data transmission {over wireless channels by adaptively modifying} the encoding rate of semantic symbols. {Initially}, {this approach conducts} feature compression based on semantic importance, {thus effectively reducing transmission overhead.}
	Subsequently, {by utilizing} real-time channel state information, the encoding rate is dynamically {adjusted} to {ensure} optimal semantic transmission. This strategy not only {enhances} the {robustness} of SemCom but also optimizes network resource utilization, {thereby facilitating} critical communications by {reducing} delays.

	\section{Potential Applications of LAM-Based MTSC} 
	Utilizing the synergy between advanced AI capabilities and next-generation communication paradigms, {the proposed LAM-based MTSC is expected to implement an intelligent communication system.}
	Some potential applications are given as follows:
	\begin{itemize}
		\item 
		\textbf{Intelligent transportation and autonomous driving:}
		By harnessing the powerful multi-modal fusion capability, the proposed LAM-based MTSC can enable vehicles to communicate with each other and their surrounding environment in a more intelligent and efficient manner.
		This technology is capable of processing vast amounts of information from diverse data sources such as sensors and cameras that provide real-time information on traffic conditions, road hazards, and pedestrian movements.
		
		\item 
		\textbf{Smart home and indoor mobile robots:}
		The proposed LAM-based MTSC can help smart home devices achieve collaborative control based on natural language, such as auto-adjusting environmental settings and offering personalized services based on user preferences, thus greatly enhancing users' living experience. 
		For mobile robots, such as cleaning or assistive robots, the proposed LAM-based MTSC can enable effective navigation, object recognition and voice command understanding in noisy environments, and even the prediction of homeowners' needs based on their behavior patterns.
		
		\item 
		\textbf{Virtual reality and immersive communications:}
		By understanding user intentions and requirements, the proposed LAM-based MTSC can employ advanced data compression techniques to reduce transmission latency in augmented reality (AR), virtual reality (VR) and mixed reality (MR), thus accelerating device response to users' actions and feedback.
		For example, in immersive communication, it has the capability to breathe new life into historical sites through interactive storytelling
		
		\item 
		\textbf{Telemedicine and remote health monitoring:}
		The proposed LAM-Based MTSC can process and analyze a multitude of health-related data streams, including vital signs, medical images, and even patients' verbal descriptions of symptoms, extracting semantic meaning to provide accurate diagnoses and personalized treatment recommendations.
		For remote medical monitoring, the proposed LAM-Based MTSC can be used to achieve low-latency transmission of high-definition image and video, thus promoting telemedicine services to a higher level of timeliness and intelligence.
	\end{itemize}

	\section{Simulation Results}
	In this section, simulation results are provided to validate the effectiveness of the proposed LAM-based MTSC scheme across multiple scenarios.
	Specifically, an encoder-decoder LAM is adopted to handle the multi-modal data after multi-task fine-tuning.
	The initial model parameters are sourced from the BART-base model.
	{Then, well-known public datasets are used to evaluate the proposed scheme's performance, including VQAv2 for the VQA task, COCO for image captioning, Audiocaps for audio captioning, MSRVTT for video captioning, and CIFAR-10 for image reconstruction.}
	In the VQA task with multi-modal input, the decoder needs to generate an answer given the semantics extracted from an image and a question.
	In the multi-task case, the decoder needs to generate different texts simultaneously to describe the input images, audio and video respectively.
	To simulate radio environments, an additive white Gaussian noise (AWGN) channel with Rician fading is considered.
	Comparative analysis is conducted against the following two baselines.
	\begin{itemize}
		\item 
		\textbf{Baseline 1 (LAM without SemCom):} The transmitter employs traditional communication techniques to perform the source and channel coding separately. The receiver uses a pre-trained multi-modal large model proposed in \cite{shukor2023unival} to complete downstream tasks. 
		\item 
		\textbf{Baseline 2 (SemCom without LAM):} The U-DeepSC framework proposed in \cite{Zhang2024A} is considered. A multi-modal semantic encoder and a channel encoder are deployed at the transmitter. A channel decoder and a unified semantic decoder are deployed at the receiver.
	\end{itemize} 

	\begin{figure}[t]
		\vspace{-5 mm}
		\centering
		{\subfloat[SemCom for the VQA task]{
				\includegraphics[width=.234\textwidth]{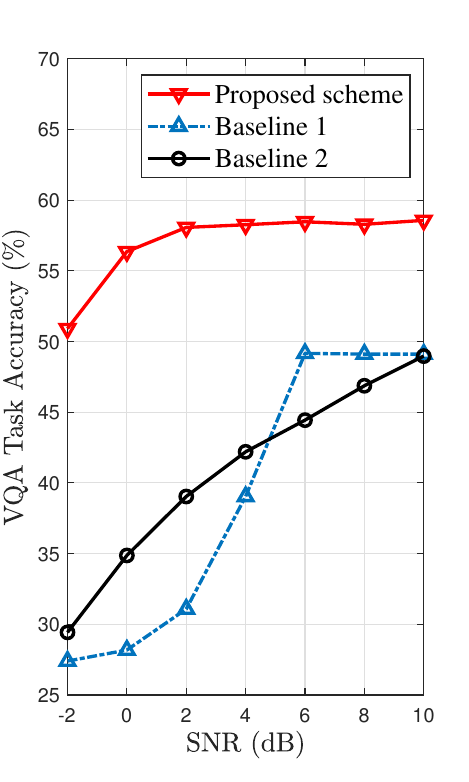}\label{Fig5_1}}}
		{\subfloat[SemCom for the captioning task]{
				\includegraphics[width=.234\textwidth]{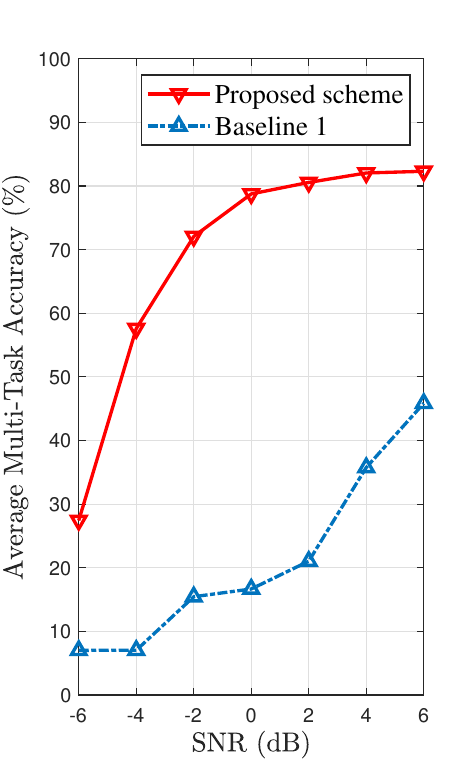}\label{Fig5_2}}}
		{\caption{Performance evaluation on the VQA and captioning tasks.}
		\label{Fig5}}
		\vspace{-4mm}
	\end{figure}
	
	\begin{figure}[t]
		\centering
		\includegraphics[width=3.5 in]{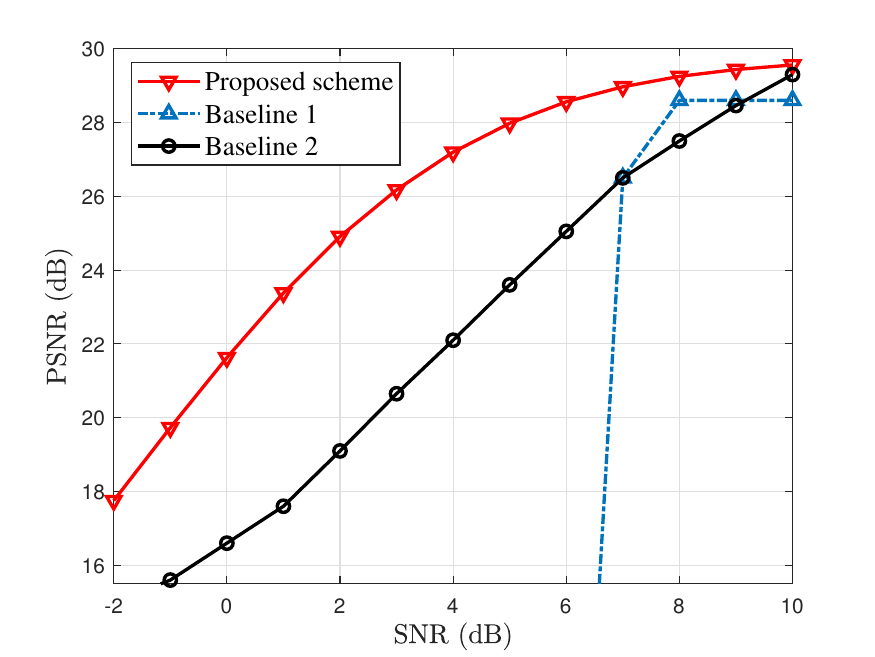}
		\caption{Performance evaluation on the image reconstruction task.}
		\label{Fig6}
	\end{figure}

	Our simulation results in Fig. \ref{Fig5} provide insights into the superiority and robustness of the proposed scheme under varying signal-to-noise ratio (SNR) conditions in the multi-modal and multi-task SemCom cases.
	Specifically, in Fig. \ref{Fig5_1}, the performance of the proposed scheme and baselines in VQA tasks is demonstrated under the varying SNR conditions, where the bilingual evaluation understudy (BLEU) score-based answer accuracy is adopted as the performance metric.
	From this figure, it can be observed that the LAM-based MTSC scheme outperforms the other two baselines, especially when the SNR is low.
	This suggests that the text generated by the model is closer to the true answer, and the LAM-based MTSC scheme is more robust against wireless noise compared to  baselines.
	Moreover, we can see that, as SNR increases to 6 dB, the transmission accuracy of image and text using traditional communication is good enough to no longer affect the inference results of the LAM at the receiver, so the performance of Baseline 1 remains stable when SNR is greater than 6 dB.
	%
	Fig. \ref{Fig5_2} illustrates the average accuracy obtained in the multi-task setup, where the receiver needs to conduct image, audio, and video captioning at the same time.
	Similar to \cite{shukor2023unival}, the consensus-based image description evaluation (CIDEr) metric is used to evaluate the quality of the generated text.
	As seen from this figure, both methods exhibit an increasing trend in average multi-task accuracy as the SNR increases, indicating that higher signal quality leads to better performance.
	Moreover, the proposed scheme consistently outperforms the baseline across all SNRs.
	In particular, when SNR is low, below 0 dB, the performance advantage of the proposed scheme is obvious, and there is a large gap with Baseline 1. At high SNR, above 0 dB, although the proposed scheme still maintains the lead, the gap between the two gradually decreases.
	{
	Fig. \ref{Fig6} plots the performance of the proposed scheme on the image reconstruction task. 
	Similar to \cite{Zhang2024A}, the peak signal-to-noise ratio (PSNR) is used to measure the quality of reconstructed images. 
	From Fig. \ref{Fig6}, we can observe that the PSNR of all schemes exhibit an upward trend as the SNR increases. Furthermore, the proposed scheme consistently outperforms other baselines across all SNR levels. This is attributed to the remarkable generalization abilities of the pre-trained LAM.}
	In summary, these observations in the above figures suggest that the proposed scheme is particularly robust against noise and performs well across multiple tasks even under challenging channel conditions.

	\section{Conclusions and Future Directions}
	In this article, the efficient model deployment and fine-tuning problems were investigated when exploiting LAMs to achieve MTSC at the network edge.
	{First of all, a LAM-based MTSC architecture was designed, including key components such as modality encoders, pre-trained LAM encoder and decoder, as well as task decoders.}
	Next, adaptive model compression and federated split fine-tuning methodologies were proposed to {adapt semantic models} to different tasks in resource-constrained wireless networks.
	In addition, a RAG scheme was presented to enhance the quality of semantic extraction and content generation {at both transmitter and receiver}.
	Moreover, several potential applications of the proposed LAM-based MTSC were {envisioned}, which aimed to integrate physical and digital worlds to improve people's daily lives.
	Finally, experimental results demonstrated that the proposed architecture can effectively extract and convey multi-modal {semantic information} across diverse tasks, thus elevating the accuracy performance under varying channel conditions.
	{Some future research directions for SemCom and large models are outlined} as follows.
	\begin{itemize}
		\item 
		\textbf{Mathematical theory of SemCom:}
		There is still a lack of semantic information theory, especially in the mathematical expression of semantics and the limit bounds of semantic compression.
		Moreover, it is important to explore a unified mathematical form to achieve accurate representation and quantitative evaluation of the semantics for multi-modal data.
		\item 
		\textbf{Balance between creativity and semantic accuracy:}
		Although large models can output high-quality contents, existing models may generate creative yet semantically incorrect or misleading content.
		To this end, it is necessary to develop new metrics or methods that can guide these large models to create highly relevant and semantic accurate contents.
		\item
		\textbf{Holistic optimization of algorithm and hardware:}
		Typically, existing algorithms and hardware are designed independently. An interesting direction involves developing innovative algorithms tailored to hardware, while utilizing dedicated accelerators to boost inference speed.
	\end{itemize}
	
	\section*{Acknowledgement}
	The work of Zhijin Qin is supported by the National Key Research and Development Program of China under Grant 2023YFB2904300 and in part by the National Natural Science Foundation of China under Grant 62293484 and 61925105.
	The work of Wanli Ni is supported in part by the Postdoctoral Fellowship Program of CPSF under Grant Number GZB20240386, and in part by the China Postdoctoral Science Foundation under Grant Number 2024M761669.
	
	\bibliographystyle{IEEEtran}
	\bibliography{IEEEabrv,reference}
	
\end{document}